\pdfoutput=1

\documentclass[11pt]{article}

\usepackage{EMNLP2023}

\usepackage{times}
\usepackage{latexsym}
\usepackage{soul}
\usepackage{lipsum}
\usepackage{graphicx}
\usepackage{xcolor}
\usepackage{stfloats}
\usepackage{lipsum}
\usepackage{amsmath}
\usepackage{animate}
\usepackage{subfigure}
\usepackage{parskip}
\usepackage[T1]{fontenc}

\usepackage[utf8]{inputenc}

\usepackage{microtype}

\usepackage{inconsolata}
\usepackage{multirow}

\newcommand{\cb}[1]{\textcolor{black}{#1}}
\usepackage[export]{adjustbox}

\title{WordArt Designer: User-Driven Artistic Typography Synthesis using \\Large Language Models}

\author{
Jun-Yan He\textsuperscript{\rm{1}}\thanks{\ \ Project Leader, Alibaba Team} \quad 
Zhi-Qi Cheng\textsuperscript{\rm{2}}\thanks{\ \ Co-Lead, CMU/RCA/ICL Team} \quad 
Chenyang Li\textsuperscript{\rm{1}} \quad 
Jingdong Sun\textsuperscript{\rm{2}} \quad 
Wangmeng Xiang\textsuperscript{\rm{1}} \\
\textbf{Xianhui Lin}\textsuperscript{\rm{1}} \quad 
\textbf{Xiaoyang Kang}\textsuperscript{\rm{1}} \quad 
\textbf{Zengke Jin}\textsuperscript{\rm{3,4}} \quad 
\textbf{Yusen Hu}\textsuperscript{\rm{2,5}} \quad 
\textbf{Bin Luo}\textsuperscript{\rm{1}} \\
\textbf{Yifeng Geng}\textsuperscript{\rm{1}} \quad 
\textbf{Xuansong Xie}\textsuperscript{\rm{1}} \quad 
\textbf{Jingren Zhou}\textsuperscript{\rm{1}} \\
\textsuperscript{1}Alibaba DAMO Academy \
\textsuperscript{2}Carnegie Mellon University \
\textsuperscript{3}Zhejiang Sci-Tech University \\
\textsuperscript{4}Royal College of Art \
\textsuperscript{5}Imperial College London\
}

\begin{document}
\maketitle

\begin{abstract}
This paper introduces \textit{WordArt Designer}, a user-driven framework for artistic typography synthesis, relying on the Large Language Model (LLM). The system incorporates four key modules: the \textit{LLM Engine}, \textit{SemTypo}, \textit{\cb{StyTypo}}, and \textit{\cb{TexTypo}} modules.~1)~The \textit{LLM Engine}, empowered by the LLM (e.g.~GPT-3.5), interprets user inputs and generates actionable prompts for the other modules, thereby transforming abstract concepts into tangible designs.~2)~The \textit{SemTypo module} optimizes font designs using semantic concepts, striking a balance between artistic transformation and readability.~3)~Building on the semantic layout provided by the \textit{SemTypo module}, the \textit{\cb{StyTypo} module} creates smooth, refined images.~4)~The \textit{\cb{TexTypo} module} further enhances the design's aesthetics through texture rendering, enabling the generation of inventive textured fonts. \cb{Notably, \textit{WordArt Designer} highlights the fusion of generative AI with artistic typography. Experience its capabilities on ModelScope: \url{https://www.modelscope.cn/studios/WordArt/WordArt}.}
\end{abstract}

\begin{figure}
\centering
\includegraphics[width=0.9\linewidth]{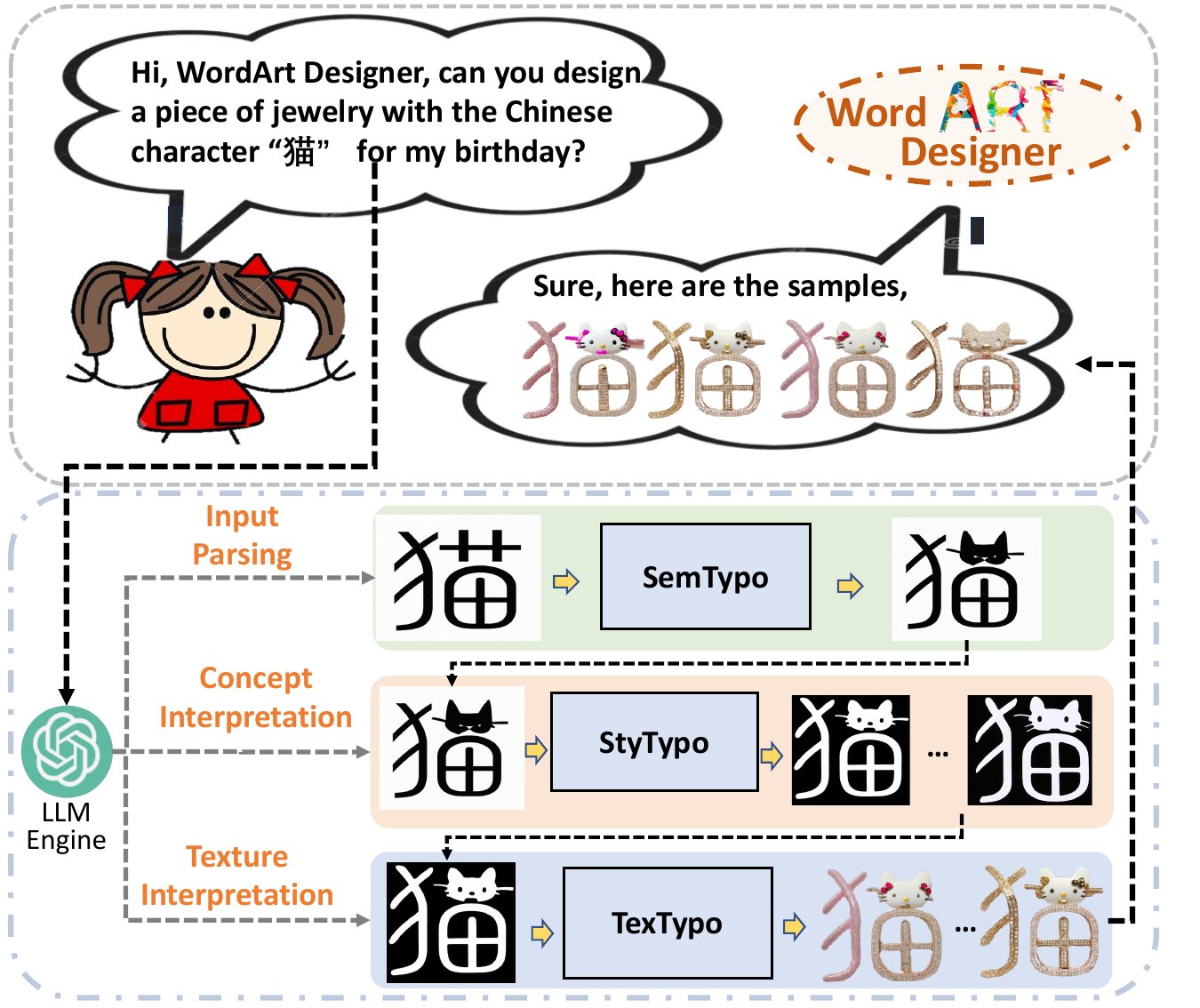}
\vspace{-2mm}
\caption{Demonstration of WordArt Designer: Leveraging the power of the LLM (e.g.~GPT-3.5), it integrates four modules (LLM Engine, SemTypo, \cb{StyTypo}, \cb{TexTypo}) to transform user inputs into visually striking and semantically rich multilingual typographic designs. It democratizes the art of typography design, enabling non-professionals to realize their creative visions.}
\label{fig:1}
\vspace{-4mm}
\end{figure}

\begin{figure*}[ht]
\begin{center}
\includegraphics[width=1.0\linewidth]{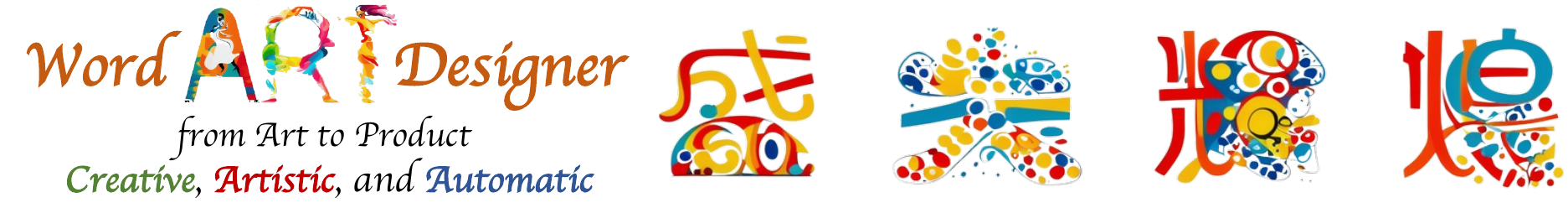}
\end{center}
\vspace{-2mm}
\caption{Examples of artistic typography generated by WordArt Designer. These instances demonstrate the system's ability to produce aesthetically pleasing, semantically coherent, and stylistically diverse typographic designs.}
\label{fig:introduction}
\vspace{-4mm}
\end{figure*}

\section{Introduction}
Typography, a critical intersection of language and design, finds extensive applications across various domains like advertising \cite{cheng2016video,cheng2017video,cheng2017video2shop,sun2018personalized}, early childhood education \cite{Vungthong2017}, and historical tourism \cite{AMAR201777}. Despite its widespread relevance, the mastery of typography design remains an intricate task for non-professional designers. Although attempts have been made to bridge this gap between amateur designers and typography \cite{Iluz_wordimage_siggraph23, Tanveer_dsfusion_corr23}, existing solutions mainly generate semantically coherent and visually pleasing typography within predefined concepts. These solutions often lack adaptability, creativity, and computational efficiency. 

To overcome these limitations, we introduce WordArt Designer (Fig.~\ref{fig:1}), a system composed of \cb{four} primary modules: the LLM Engine, SemTypo Module, and \cb{StyTypo} Module, supplemented by the \cb{TexTypo} Module for texture rendering. This user-focused system allows users to define their design needs, including design concepts and domains. The system consists of:

\begin{enumerate}
    \item \textbf{LLM Engine}: Based on the power of the LLM (e.g.~GPT-3.5), this engine interprets user input and produces prompts for the SemTypo, \cb{StyTypo}, and \cb{TexTypo} modules.
    \item \textbf{SemTypo Module}: The SemTypo Module transforms typography based on a provided semantic concept. It involves a three-step process, including Character Extraction and Parameterization, Region Selection for Transformation, and Semantic Transformation \& Differentiable Rasterization.
    \item \textbf{\cb{StyTypo} Module}: The \cb{StyTypo} Module generates smoother, more detailed images based on the semantic layout image provided by the SemTypo module.
    \item \textbf{\cb{TexTypo} Module}: The \cb{TexTypo} Module modifies ControlNet for texture rendering, ensuring creativity while preserving legibility. 
\end{enumerate}

The workflow, illustrated in \cb{Fig.~\ref{fig:1}}, begins with the LLM module interpreting user input. The output of each module serves as the input for the next, with the final design decision made by the \cb{TexTypo} module. This sequence ensures the final design aligns with the user's intent and maintains a unique aesthetic appeal.

This design process is iterative, involving constant interaction between the user's input and the system's modules. This user-centered approach guarantees the creation of high-quality WordArt designs \cb{(See Fig.~\ref{fig:introduction})}, making it an effective tool in creative design-dependent industries, such as food and jewelry.

Extensive experiments on WordArt Designer have validated its creativity, artistic expression, and expandability to different languages. The inclusion of a ranking model significantly improves the success rate and overall quality of stylized images, ensuring the production of high-quality WordArt designs.

In essence, WordArt Designer provides a creative, artistic, and fully automated solution for generating word art. Our research not only lays the groundwork for future text synthesis studies but also introduces numerous practical applications. WordArt Designer can be employed in various areas, including media propaganda and product design, enhancing the efficiency and effectiveness of these systems, thereby making them more practical for everyday use.

\section{Related work}
\noindent \textbf{LLM and their Apps}.~Large Language Model (LLM) has been progressively improved and utilized in a wide range of applications \cite{Anil_PaLM_Corr19, Raffel_T5_JMLR20, Shoeybi_MegatronLM_Corr19, Rajbhandari_ZeRO_SC20, Devlin_BERT_NAACLHLT19,cheng2023chartreader}. The outstanding performances exhibited by the ChatGPT and GPT series \cite{Radford_Improving_OPENAI18, Brown_GPT3_NIPS20, OpenAI_GPT4_Corr23} have stimulated the widespread use of the LLM. These models are adept at learning context from simple prompts, leading to their increasing use as the controlling component in intelligent systems \cite{Wu_VisualChatGPT_Corr23, shen_hugginggpt_Corr23}. Building on these insights, WordArt Designer incorporates the LLM to enhance system creativity and diversity.

\noindent \textbf{Text Synthesis}.~While significant progress has been made in image synthesis, integrating legible text into images remains challenging \cite{Rombach_LDM_CVPR22, Saharia_Imagen_NIPS22}. Some solutions, such as eDiff-I \cite{Balaji_eDiffiI_Corr22} and DeepFloyd \cite{DeepFloyd_23}, employ robust LLMs, such as T5 \cite{Raffel_T5_JMLR20}, for improved visual text generation. Recent studies \cite{Yang_GlyphControl_Corr23, Ma_glyphdraw_Corr23} have also looked into using glyph images as extra control conditions, while others like DS-Fusion \cite{Tanveer_dsfusion_corr23} introduce additional constraints to synthesize more complex text forms, such as hieroglyphics.

\noindent \textbf{Image Synthesis}.~The surge in demand for personalized image synthesis has spurred advances in interactive image editing \cite{Meng_SDEdit_ICLR22, Gal_Inversion_ICLR23, Brooks_InstructPix_Corr22,zhao2018multi} and techniques incorporating additional conditions, such as masks and depth maps \cite{Rombach_LDM_CVPR22,huang2020generating}. New research \cite{Zhang_ControlNet_Corr23, Mou_T2I_Corr23, Huang_Composer_Corr23} is exploring multi-condition controllable synthesis. For instance, ControlNet \cite{Zhang_ControlNet_Corr23} learns task-specific conditions end-to-end, providing more nuanced control over the synthesis process.

\noindent \textbf{Text-to-Image Synthesis}.~Significant strides in denoising diffusion probabilistic models have substantially enhanced text-to-image synthesis \cite{Ho_DDPM_NIPS20, Ramesh_DALLE_icml21, Song_SDE_ICLR21, Dhariwal_GuidedDiffusion_NIPS21, Nichol_ImprovedDDPM_ICML21, Saharia_Imagen_NIPS22, Ramesh_DALLE2_corr22, Rombach_LDM_CVPR22}. Notable examples of these advancements are latent diffusion models such as Imagen \cite{Saharia_Imagen_NIPS22}, DALLE-2 \cite{Ramesh_DALLE2_corr22} and LDM \cite{Rombach_LDM_CVPR22}, which have enabled high-quality image generation.

\begin{figure*}[!ht]
\begin{center}
\includegraphics[width=\linewidth]{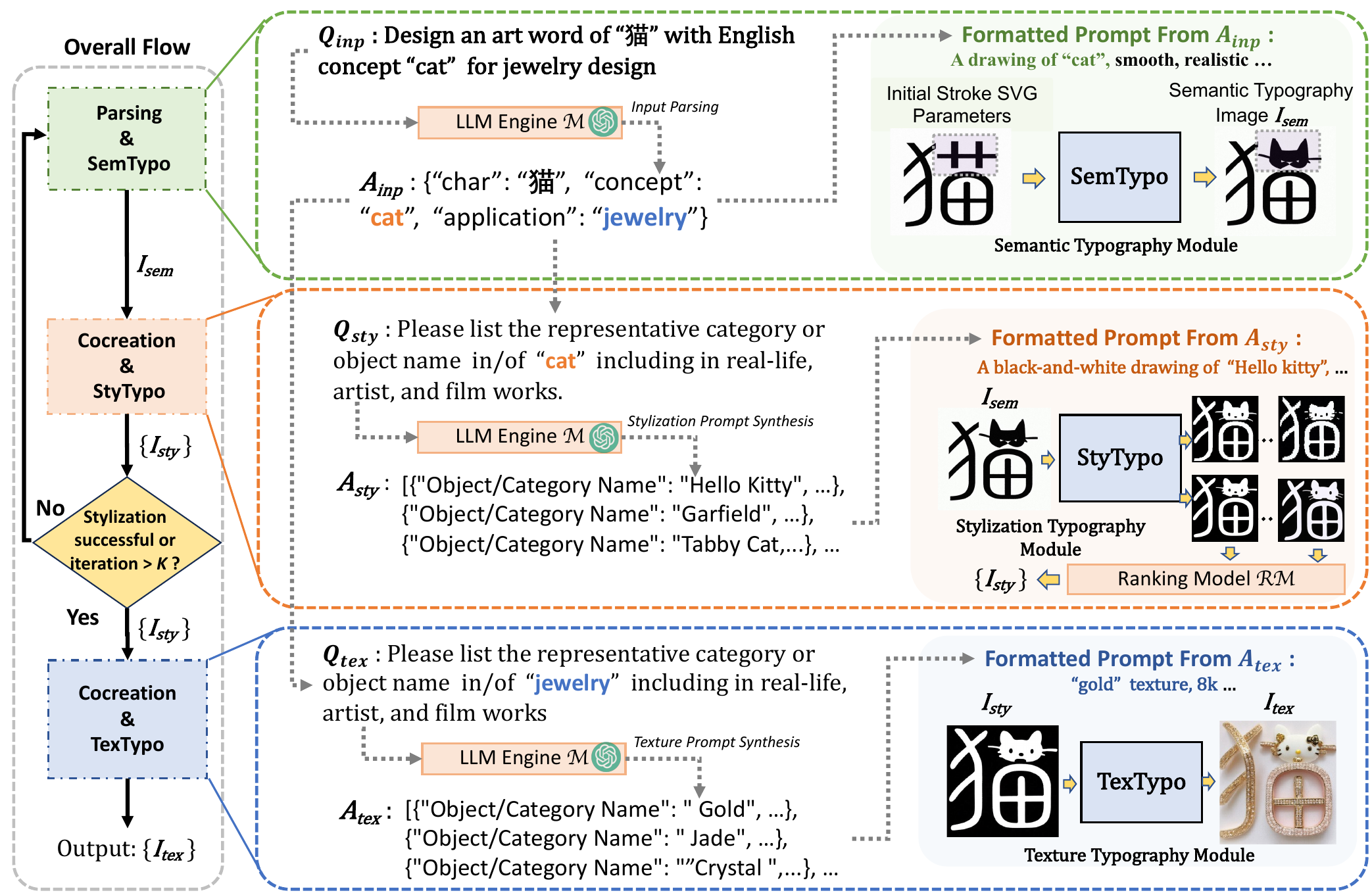}
\end{center}
\vspace{-2mm}
\caption{The architectural framework of the proposed WordArt Designer system. This structure involves an LLM engine, the SemTypo module for Semantic Typography, the \cb{StyTypo} module for Stylization Typography, and the \cb{TexTypo} module for Texture Typography. These modules operate coherently, guided by a preset control flow, to facilitate a seamless and innovative transformation of text into artistic typography.}
\label{fig:framework}
\vspace{-4mm}
\end{figure*}

\section{WordArt Designer}
The WordArt Designer system utilizes an assortment of typography synthesis modules, propelled by a Large Language Model (LLM) such as GPT-3.5), facilitating an interactive, user-centered design process. As illustrated in Fig.~\ref{fig:framework}, users define their design needs, including design concepts and domains, e.g., "A cat in jewelry design." The LLM engine interprets the input, generating prompts to guide SemTypo, \cb{StyTypo}, and \cb{TexTypo} modules, thus executing the user's design vision.

To achieve automated WordArt design, we introduce a quality assessment feedback mechanism, which is vital for successful synthesis.
The output from the ranking model is evaluated by the LLM engine to validate the quality of the synthesized image, ensuring the creation of at least $K$ qualified transformations. 
If this criterion is not met, the LLM engine, along with the SemTypo and \cb{StyTypo} modules and format directives, are restarted for another design iteration. Subsequent sections will delve into the details of each module's functionality and operation.

\subsection{LLM Engine}
The Large Language Model (LLM) engine is a crucial component for the WordArt designer. It serves as a knowledge engine and concretizes abstract notions, like "vegetables" and "fruit", into texture prompts in the context of food, for the eventual synthesis of the artistic text. For most concrete nouns, such as "cat", "dog", "flower", etc., semantic typography can be successfully generated. However, for words like abstract nouns and verbs, such as "winter", "hit", etc., users often fail to provide desired descriptions. The reason is that images compose highly complex scenes for abstract concepts, which is not suitable for our WordArt designer system.

To address this issue, we employ the LLM to render abstract concepts into representative objects that can be easily converted. \cb{Specifically, we can build our LLM engine using models like GPT-3.5 and other LLMs, all of which have context-learning capabilities.} The prompts for input parsing, stylization, and texture rendering are generated as:
\begin{equation}
\small
A_{inp} = \mathcal{M}(Q_{inp}),
A_{sty} = \mathcal{M}(Q_{sty}), \\
A_{tex} = \mathcal{M}(Q_{tex}) 
\end{equation}
Where $Q_{inp}$, $Q_{sty}$, and $Q_{tex}$ represent the standard prompts for input parsing, stylization, and texture rendering respectively. $Q_{sty}$ and $Q_{tex}$ are built using formatted prompt templates with concepts derived from the input parsing. LLM engine has ample capabilities to imbue our system with a creative and engaging "soul", ensuring the quality of artistic text synthesis. We provide detailed templates and full examples of prompts in Appendix~\ref{prompt-examples}.

\subsection{SemTypo Module}
The Semantic Typography (SemTypo) module alters typographies based on a given semantic concept. It unfolds in three stages: (1) Character Extraction and Parameterization, (2) Region Selection for Transformation, and (3) Semantic Transformation and Differentiable Rasterization.

\noindent \textbf{Character Parameterization}.~The first stage, as displayed in Fig.~\ref{fig:framework}, starts by transforming the natural language input into a JSON format, specifying the characters to modify, the semantic concept, and the application domain. The FreeType font library~\cite{david_turner_freetype_1996} is then employed to extract character contours and convert them into cubic Bézier curves characterized by a trainable set of parameters. For characters with surplus control points, a subdivision routine fine-tunes the control points $\theta$, using a differentiable vector graphic rasterization scheme~\cite{Iluz_wordimage_siggraph23}.

\noindent \textbf{Region Selection}.~Our unique contribution is the region-based transformation method, the second stage of the SemTypo module. This approach facilitates the selective transformation of certain character segments, effectively reducing distortions that typically affect typography generation in languages with single-character words. We choose to transform a random contiguous subset of control points within a character, instead of the entire character. We establish a splitting threshold of 20 pixels, with the set of control points randomly determined within the range [500, min(1000/control point count)], initiating from a random point.

In contrast to previous methods, such as the one by Iluz et al.~\cite{Iluz_wordimage_siggraph23}, which used extra loss terms with inadequate success to maintain legibility of the synthesized typography, our method only involves loss computation from the transformed sections of the characters. This approach increases efficiency and guarantees careful manipulation of character shapes, thus improving transformation quality.

\begin{figure*}[ht]
\begin{center}
\includegraphics[width=0.95\linewidth]{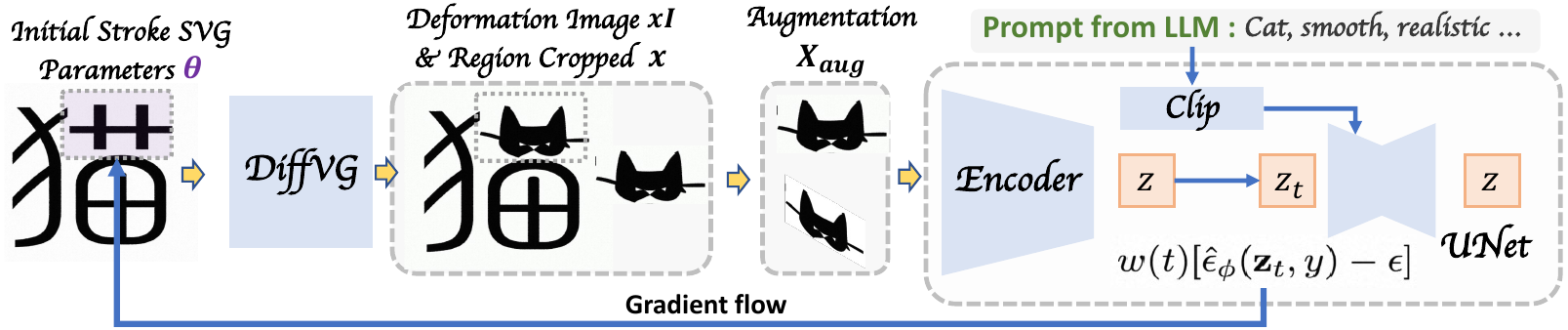}
\end{center}
\vspace{-2mm}
\caption{\cb{Differential} rasterization scheme of semantic typography. \cb{The character stroke inside the \textcolor{purple}{purple box} is the selected part for optimization.} }
\label{fig:rasterization}
\vspace{-4mm}
\end{figure*}

\noindent \textbf{Transformation and Rasterization}.~In the final stage, the parameters are transformed and rasterized through the Differentiable Vector Graphics (DiffVG) scheme~\cite{Li_DiffVG_SIGGraph20}. As shown in Fig.~\ref{fig:rasterization}, the transformed glyph image $I_{sem}$ is created from the trainable parameters $\theta$ of the SVG-format character input, using DiffVG $\boldsymbol{\phi}(\cdot)$. A segment of the chosen character $x$ is optimized and cropped to yield an enhanced image batch $X_{aug}$~\cite{Frans_Clipdraw_NIPS22}. The semantic concept $S$ and the augmented image batch $X_{aug}$ are both input into a vision-language backbone model to compute the loss for parameter optimization. The Score Distillation Sampling (SDS) loss is applied in the latent space code $z$, as per the DreamFusion method~\cite{Poole_Dreamfusion_ICLR23}:
\begin{equation}
\centering \small
\nabla_{\theta} \mathcal{L}_{SDS} = \mathcal{E}_{t,\epsilon}[w(t) (\hat{\epsilon_{\phi}}(a_{t}z_{t}+\sigma_{t}\epsilon, y)-\epsilon) \frac{\partial z}{\partial X_{aug}} \frac{\partial X_{aug}}{\partial \theta}]
\end{equation}
Here, $t\in \{1,2, \dots, T\}$ is uniformly sampled to define a noise latent code $z_{t} = a_{t}z_{t}+\sigma_{t}\epsilon$, with $\epsilon N \sim (0, 1)$, and $a_{t}$, $\sigma_{t}$ act as noise schedule regulators at time $t$. The multiplier $w(t)$ is a constant, contingent on $a_{t}$. This revised process refines expression and amplifies the variety of output.

\subsection{\cb{StyTypo} Module}
The Stylization Typography (\cb{StyTypo}) module's main purpose is to generate smoother and more detailed images, enhancing the semantic layout image $I_{sem}$. To speed up $I_{sty}$ generation, we use short iteration settings. However, this approach might lead to a lack of smoothness and details. To overcome these potential drawbacks, the \cb{StyTypo} module introduces two main components: (1) stylized image generation, and (2) stylized image ranking and selection.

\begin{figure}[ht]
\begin{center}
\includegraphics[width=0.95\linewidth]{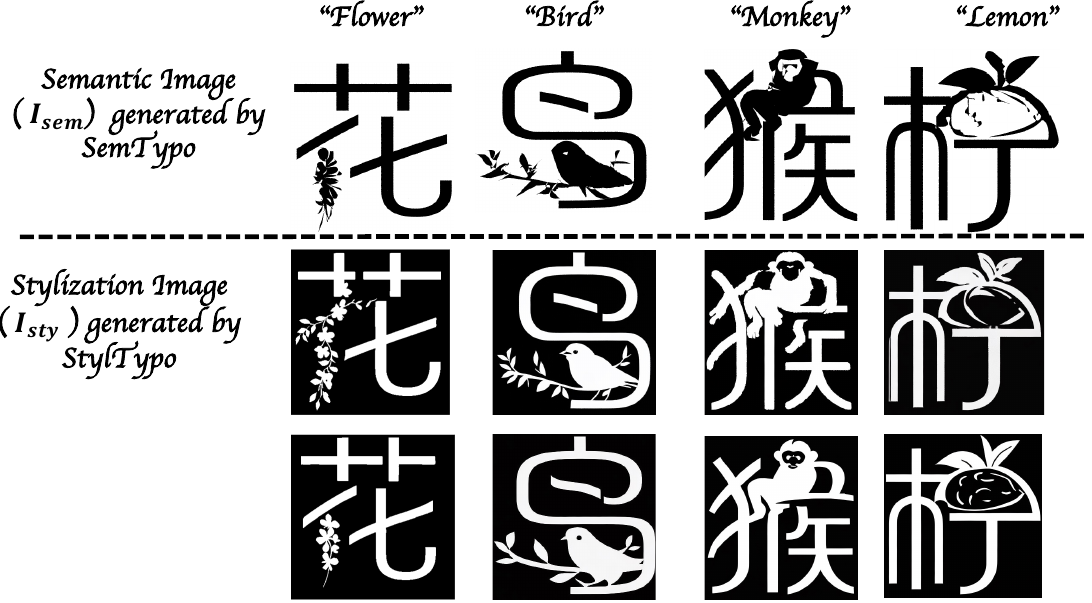}
\end{center}
\vspace{-1mm}
\caption{Comparison of the semantic and stylization images. Stylization images contain more details.}
\label{fig:styl-case}
\vspace{-2mm}
\end{figure}

\noindent \textbf{Stylized Images Generation}.~The Latent Diffusion Model (LDM)~\cite{Rombach_LDM_CVPR22} has gained attention for its ability to generate images based on given input shapes. Therefore, we employ the LDM's depth2image methodology to stylize typographic layouts, enhancing smoothness and infusing additional detail to create a unique "sketch" for texture rendering. Fig.~\ref{fig:styl-case} illustrates this, where the top row images generated by the SemTypo module, despite lacking smoothness and detail, provide a comprehensive object representation. After being processed by the \cb{StyTypo} module, the stylized images on the lower row display an abundance of detail and inventive renderings for each semantic image input.

Formally, given a semantic typography image $I_{sem}$ from the SemTypo module, and a stylization prompt $A_{sty}$ synthesized by the LLM Engine $\mathcal{M}$, we can create the stylization image $I_{sty}$ as:
\begin{equation}
    I_{sty} = \text{\cb{StyTypo}}(I_{sem}, A_{sty})
\end{equation}
where \cb{StyTypo} utilizes the depth2image pipeline derived from the LDM~\cite{Rombach_LDM_CVPR22} to carry out the stylization. 

\noindent \textbf{Ranking and Selection}.~To augment the \cb{StyTypo} module's efficiency, we introduce a ranking model that orders and filters the generated results. Specifically, we establish a quality evaluation dataset consisting of stylized characters classified into two groups: (1) Successful Stylization, and (2) Failed Stylization. The dataset encompasses 141 single-character Chinese characters and 5,814 stylized typographic images. We leverage the ResNet18 classification model~\cite{DBLP:conf/cvpr/HeZRS16} to learn the quality distribution of the stylization images. During the filtering stage, the trained model serves as a ranking model, providing ranking scores. Based on these scores, the top 'x' results are selected.

\subsection{\cb{TexTypo} Module}
To advance the styling capacities of the Stylization Typography (\cb{StyTypo}) module, we adapted ControlNet~\cite{Zhang_ControlNet_Corr23} for the purpose of texture rendering, resulting in the creation of the Texture Typography (\cb{TexTypo}) module.

\begin{figure}[ht]
\begin{center}
\includegraphics[width=0.85\linewidth]{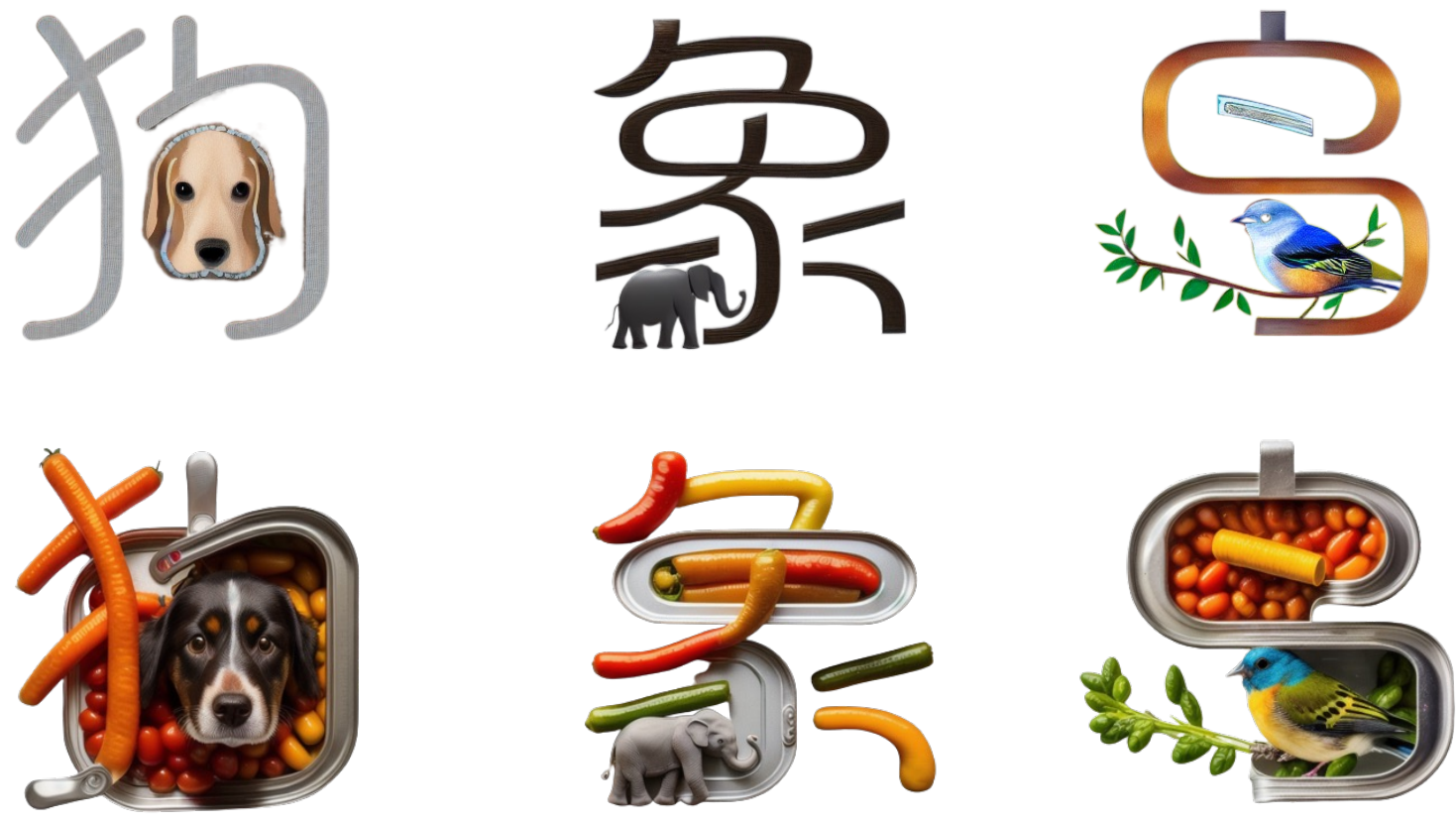}
\end{center}
\vspace{-1mm}
\caption{Comparison between Canny Edge and Scribble conditions for ControlNet texture rendering. The first row is generated using the Canny Edge condition, while the rest are from the Scribble condition. }
\label{fig:canny-scribble}
\vspace{-2mm}
\end{figure}

As can be seen in Fig.~\ref{fig:canny-scribble}, ControlNet's original control conditions relied heavily on the Canny Edge and Depth data. This constraint tended to produce fonts that were lacking in creativity and artistic flair. To counter this, we introduced Scribble conditions as an alternate control condition \cb{into ControlNet}, which encourage the generation of more creatively textured fonts without compromising on readability.

Furthermore, to cater to a range of industrial sectors, we have reconfigured ControlNet to incorporate pre-trained stable diffusion models that are relevant to different fields. These include, but are not limited to, commercial advertising, fashion design, gaming interfaces, tech products, and artistic creations.

Technically, we provide the ControlNet parameters with conditions Canny Edge, Depth, Scribble, as well as original font images. The \cb{TexTypo} model receives these parameters and generates the textured font image as,
\begin{equation}
I_{tex} =\text{\cb{TexTypo}}(I_{sty}, A_{tex}, P_{cond}),
\end{equation}
where $A_{tex}$ represents the prompts synthesized by the LLM engine $\mathcal{M}$, and $P_{cond}$ stands for the control parameters, resulting in a creatively rendered textured font as the output.

\begin{figure*}[!ht]
\begin{center}
\includegraphics[width=\linewidth]{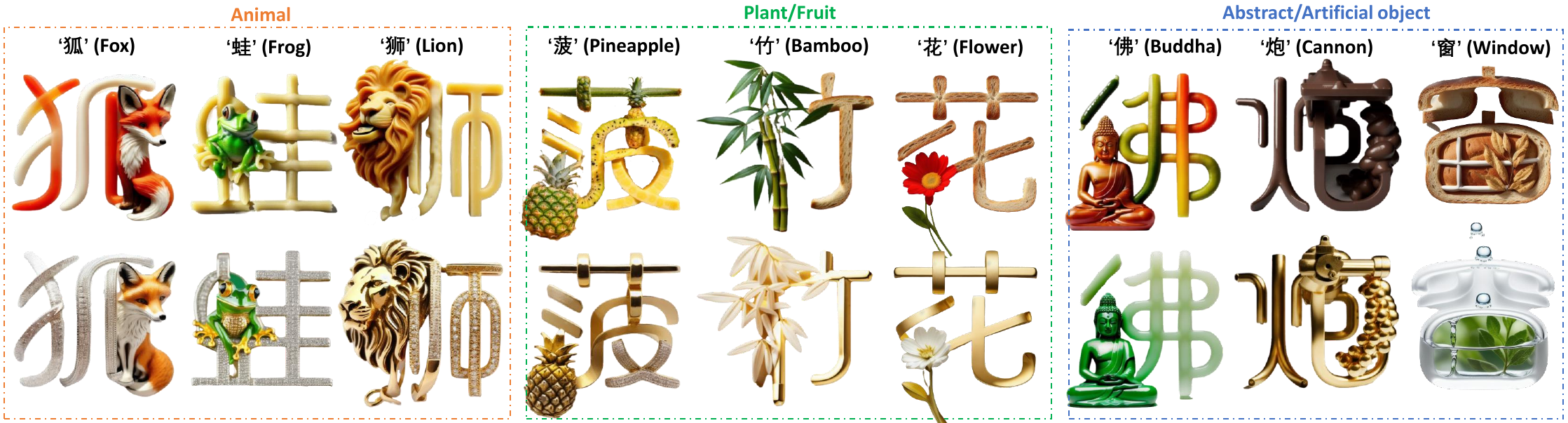}
\end{center}
\vspace{-2mm}
\caption{\cb{Results showcasing the adaptability of the WordArt Designer. The first row targets the concept of ``food'', which is further specified to ``candy'', ``pasta'', ``cheese'', ``fruits'', ``bread'', ``vegetables'' or ``chocolate''. The second row targets ``jewelry'', concretized to ``jewels'', ``gold'' or ``jade''.} The variety of styles highlighted underscores WordArt Designer's versatility in creating unique artistic typography, pushing past traditional design boundaries.}
\vspace{-2mm}
\label{fig:texttypo_llm}
\end{figure*}

\section{Deployment Details}
\cb{WordArt Designer tool has successfully been integrated into ModelScope, utilizing the TongYi QianWen 14b model as the LLM engine. In terms of deployment, the StyTypo and TexTypo Modules are hosted in Docker containers, ensuring both flexibility and scalability in deployment. StyTypo is powered by a Linux platform with 48 cores, 384GB RAM, and 4 Nvidia V100 32GB GPUs, taking roughly 32 seconds to generate 4 images. In contrast, TexTypo operates within a similar Linux environment but with 24 cores, 64GB RAM, and a single Nvidia V100 32GB GPU, and typically produces 4 images within a span of 5 to 10 seconds. For the Ranking Model, the \texttt{mmpretrain} \cite{mmpretrain2023} is used to train a ResNet 18 model \cite{DBLP:conf/cvpr/HeZRS16}, with a total of $100$ epoches at a batch size of $32$. The SGD optimizer was used with a learning rate of $0.01$. The training ran on a single Nvidia V100 16GB GPU.}

\section{Experiments}
\noindent \textbf{Creativity \& Artistic Ability}.~We operationalize the concept of texture rendering to evaluate the Creativity and Artistic Ability of the WordArt Designer. The outcomes are demonstrated in Fig.~\ref{fig:texttypo_llm}. The first row of art words is generated by embodying the concept ``food'', which is further specified to ``candy'', ``pasta'', ``cheese'', ``fruits'', ``bread'', ``vegetables'' or ``chocolate''. The second row represents the concept ``jewelry'', concretized to ``jewels'', ``gold'' or ``jade''. The smart and reasonable texture rendering contributes to the creativity and artistic appeal of the output.

\begin{figure}[ht]
\begin{center}
\includegraphics[width=\linewidth]{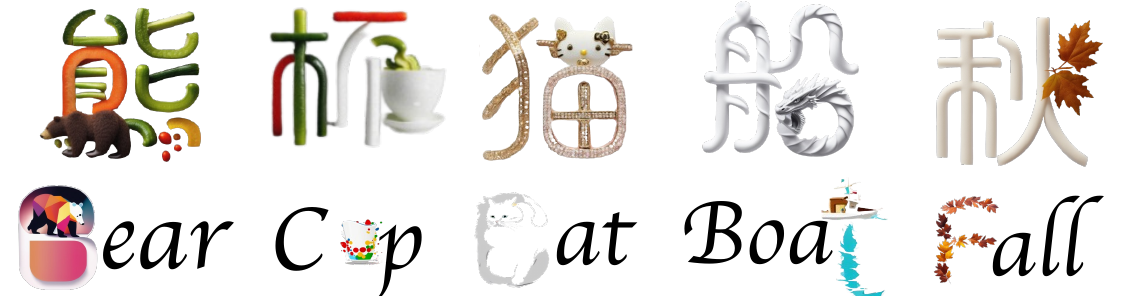}
\end{center}
\vspace{-0.1in}
\caption{Chinese Characters and their corresponding English art words.}
\label{fig:multi-lang}
\end{figure}

\noindent \textbf{Expandability to Different Languages}.~Our SemTypo module, grounded on differentiable rasterization, is theoretically compatible with all types of languages. Beyond Chinese (i.e., hieroglyphs), we explore the expandability of WordArt Designer with the representative language, English (i.e., \cb{the Latin alphabet}). Fig.~\ref{fig:multi-lang} presents a collection of rendered results for Chinese characters and corresponding English words, substantiating that WordArt Designer effectively accommodates different languages.

\cb{\noindent \textbf{Effect of Ranking Model}.~To determine the effectiveness of the ranking model, we divide the aforementioned character dataset into a training and validation set by randomly selecting 20 characters for validation. We use precision and recall to measure the model's ability to classify individual images as successfully stylized or not. In addition, we compare WordART Designer's overall success rate on transforming a character using the Ranking Model and a Random Model (a character is deemed successfully transformed if at least one of the output images is successfully stylized).} As shown in Table~\ref{table:ranking_exp}, our ranking model significantly outperforms the random model, indicating its efficacy. When top-10 images are selected, we guarantee that each character has at least one successfully stylized image. To balance precision and recall, the number of selected images should ideally range from 2 to 5.

\begin{table}
\scriptsize
\centering
\caption{Ablation study of the ranking model on the validation set. `p', `r', and `s' stand for precision, recall, and success rate, respectively. `x' in `TopX' indicates the number of stylized images retained. In the ranking-based method, `TopX' are selected based on ranking scores, while for the random-based method, `TopX' are selected randomly. Results of the random-based method are obtained by averaging over 10,000 iterations. Increased values are indicated in \textcolor[RGB]{0,102,255}{blue}.}
\setlength{\tabcolsep}{1.35mm}{
\begin{tabular}{c|c|c|c|c|c}
\hline \hline
Methods                 & Metric & Top1          & Top2          & Top5          & Top10          \\ \hline
\multirow{3}{*}{Random} & p      & 18.3          & 18.1          & 18.2          & 18.2           \\ \cline{2-6} 
                        & r      & 4.5           & 8.9           & 22.4          & 44.8           \\ \cline{2-6} 
                        & s      & 18.3          & 33.1          & 63.4          & 86.5           \\ \hline
\multirow{3}{*}{Ranking}   & p      & \textbf{60.0}\textcolor[RGB]{0,102,255}{$\uparrow$41.7} & \textbf{62.5}\textcolor[RGB]{0,102,255}{$\uparrow$44.4} & \textbf{46.0}\textcolor[RGB]{0,102,255}{$\uparrow$27.8} & \textbf{32.0}\textcolor[RGB]{0,102,255}{$\uparrow$13.8}  \\ \cline{2-6} 
                        & r      & \textbf{14.6}\textcolor[RGB]{0,102,255}{$\uparrow$10.1} & \textbf{30.8}\textcolor[RGB]{0,102,255}{$\uparrow$21.9} & \textbf{56.3}\textcolor[RGB]{0,102,255}{$\uparrow$33.9} & \textbf{78.8}\textcolor[RGB]{0,102,255}{$\uparrow$34.0}  \\ \cline{2-6} 
                        & s      & \textbf{60.0}\textcolor[RGB]{0,102,255}{$\uparrow$41.7} & \textbf{80.0}\textcolor[RGB]{0,102,255}{$\uparrow$46.9} & \textbf{85.0}\textcolor[RGB]{0,102,255}{$\uparrow$21.6} & \textbf{100.0}\textcolor[RGB]{0,102,255}{$\uparrow$13.5} \\ \hline \hline
\end{tabular}}
\label{table:ranking_exp}
\end{table}

\begin{figure}[!ht]
\begin{center}
\includegraphics[width=1.\linewidth]{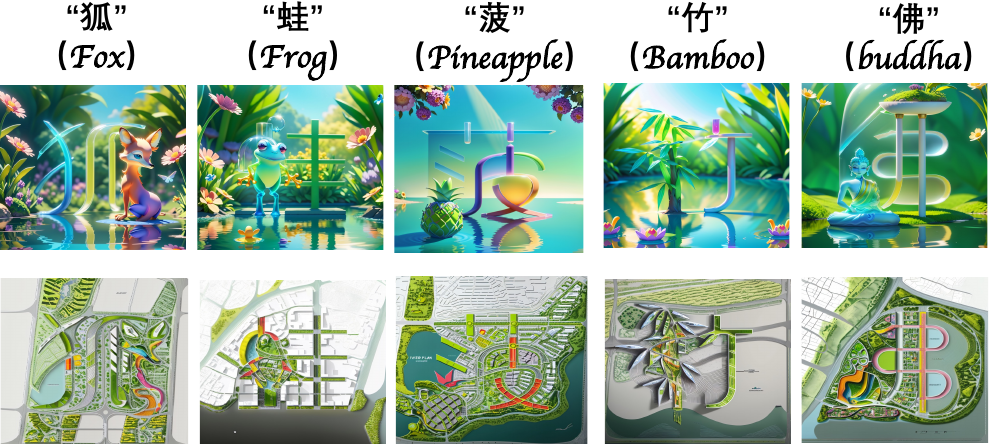}
\end{center}
\vspace{-2mm}
\caption{Various notable applications of our WordArt Designer, including art word poster creation (row 1) and urban master plan design (row 2). Note that \href{https://civitai.com/models/7371/rev-animated}{revAnimated} is used as the base LDM~\cite{Rombach_LDM_CVPR22}. For rows 1-2, we further apply the Lora models \href{https://civitai.com/models/25995/blindbox}{Blindbox} and \href{https://civitai.com/models/91751/jiangdamasterplan}{MasterPlan} respectively.}
\label{fig:texture-application}
\vspace{-2mm}
\end{figure}

\subsection{Application}
\noindent \textbf{WordArt Image}.~We experiment with various application possibilities for WordArt Designer. The results, exhibited in Fig.~\ref{fig:texture-application}, are representative and not cherry-picked. WordArt Designer exhibits promising potential in areas like art word poster design and even city planning. We are confident that WordArt Designer will bring innovative inspiration to professional designers.

\noindent \textbf{WordArt Animation}.~We also utilize ControlVideo~\cite{Zhang_ControlVideo_Corr} to synthesize art word videos, illustrating the transformation of the word/character. The Chinese characters for Bamboo'' and Flower'' are used in the video generation process, with the "Van Gogh's painting" style applied to the animations, proving useful for Chinese education. \cb{Please refer to Fig.~\ref{video:video-demo} for additional animations.}

\begin{figure}[htbp]
    \centering
    \subfigure[\scriptsize Bamboo~(van~Gogh)]{
        \animategraphics[width=.35\linewidth]{6}{videos/Vincent-van-Gogh-style-bamboo/frame-}{0}{11}
    }
    \subfigure[\scriptsize Follower~(van~Gogh)]{
        \animategraphics[width=.35\linewidth]{6}{videos/Vincent-van-Gogh-style-flower/frame-}{0}{11}
    }
\vspace{-0.15in}
    \caption{\textbf{Art word animations} derived from the SemTypo optimization process. \textcolor{orange}{CLICK the image to PLAY ANIMATION! Best viewed with Adobe Acrobat DC}.}
    \label{video:video-demo}
    \vspace{-2mm}
\end{figure}

\section{\cb{Ethical Considerations}}

\cb{Potential ethical concerns include perpetuating cultural stereotypes due to the use of certain imagery or symbols in the process of artistic transformations, or introducing bias against under-represented cultures. Another issue could be the potential inclusion of copyrighted graphics. Users need to pay attention to these issues to ensure responsible and respectful use of the system.}

\section{Conclusion}
This paper presents WordArt Designer, a framework that harnesses Large Language Models (LLM), such as  GPT-3.5, to automatically generate multilingual artistic typography. This system uses an LLM engine to parse and translates user input into directives, guiding three modules, each accountable for different aspects of the typographic design. The superior performance of WordArt Designer highlights the potential of AI to augment artistic typography. Future work aims to further explore the possibilities of integrating this technology into other aspects of design, such as graphics and interactive media. 

\section*{Acknowledgments}
The contributions of Zhi-Qi Cheng in this project were supported by the Army Research Laboratory (W911NF-17-5-0003), the Air Force Research Laboratory (FA8750-19-2-0200), the U.S. Department of Commerce, National Institute of Standards and Technology (60NANB17D156), the Intelligence Advanced Research Projects Activity (D17PC00340), and the US Department of Transportation (69A3551747111).  Intel and IBM Fellowships also provided additional support for Zhi-Qi Cheng's research work. 

{
\bibliographystyle{acl_natbib}
\bibliography{ref}
}

\clearpage

\appendix

\onecolumn

\section{Prompt Examples} \label{prompt-examples}
\noindent \textbf{Prompt for GPT}: ``\textit{Now you are a creative and active explainer that helps people to understand the abstract concept, and concrete the abstract concepts, tell them the representative object in the abstract concept. All the question is in a standard format "Please list the representative category or object name in/of <CONCEPT>, including in real-life, artist, and film works." And the output must be in a strict JSON format: \{"Object/Category Name": "Name", "description": "less than five critical words to describe", "reason": "your detailed reason for the choice"\}. Example: " Please list the representative category or object name in/of cat including in real-life, artist, and film works." The response should be \{"Object/Category Name": "Hellokitty", "description": "cute, happiness" "less than five critical words to describe", "reason": "famous for the cartoon"\}. }''

\noindent \textbf{Stylization and texture examples as follow:}

\noindent \textbf{$Q_{s}$: ``Please list the representative category or object name in/of spring, including in real-life, artist, and film works.''}

$A_{s}$: ``\textit{\{"Object/Category Name": "Rainbow", "description": "colorful, natural", "reason": "Rainbows are a natural phenomenon that occurs after rain showers during spring. They are often depicted in artwork and films as a symbol of hope, joy, and promise. Additionally, rainbows are often used in fashion and design to represent spring and its vibrant colors."\}}''

\noindent \textbf{$Q_{t}$: ``Please list the representative category or object name in/of food, including in real-life, artist, and film works.''}

$A_{t}$: ``\textit{\{"Object/Category Name": "Pizza", "description": "delicious, versatile", "reason": "Pizza is a popular food that is loved by many people around the world. It is a versatile food that can be customized with a variety of toppings to suit different tastes and preferences. Pizza is often featured in films, TV shows, and commercials, and it is a staple food in many countries, including Italy and the United States."\}}''

The ``Object/Category Name'' and the ``description'' are utilized to build the prompt for the \cb{StyTypo} and \cb{TexTypo} modules, and the ``reason'' information can be applied to analyze the quality of the prompt.

\section{Addtional Results}

\begin{figure*}[ht]
\begin{center}
\includegraphics[width=0.9\linewidth]{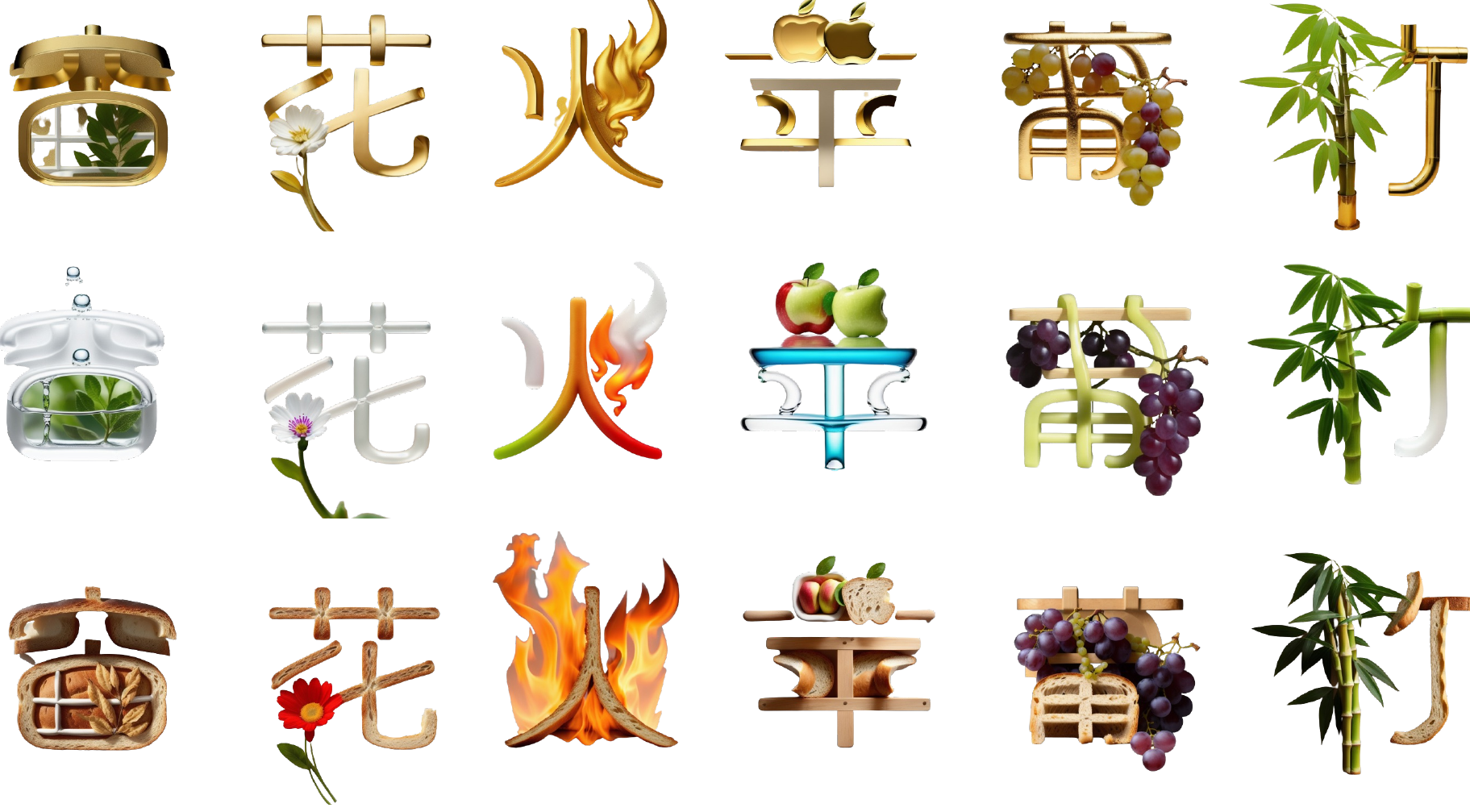}
\end{center}
\vspace{-2mm}
\caption{Diversity of results. The LLM engine generates the texture prompt that can be explained in various concretion objects/concepts. The 1st and 2nd rows are related to the concept ``jewelry'' that is concrete to ``gold'' or ``jade'', respectively.   The 3rd row is the concept ``food'' that is concrete to bread. It is worth noticing that the texture rendering is ``smart'' and ``reasonable'' which leads to creativity and artistry.}
\label{fig:texttypo}
\vspace{-2mm}
\end{figure*}

\end{document}